\documentclass[runningheads]{llncs}

\usepackage{eccv}

\usepackage{eccvabbrv}

\usepackage{graphicx}
\usepackage{booktabs}

\usepackage[accsupp]{axessibility}  

\usepackage[pagebackref,breaklinks,colorlinks,citecolor=eccvblue]{hyperref}

\usepackage{orcidlink}
\usepackage{placeins}
\usepackage{multirow}
\usepackage{booktabs}
\usepackage{caption}
\usepackage{subcaption}
\usepackage{pifont}
\usepackage{wrapfig}
\usepackage{graphicx}
\usepackage{makecell}
\usepackage{tikz}
\newcommand{\cmark}{\textcolor{green!70!black}{\ding{51}}}
\newcommand{\xmark}{\textcolor{red}{\ding{55}}}
\newcommand{\ourmethod}{\textsc{URoPE}\xspace}

\begin{document}

\title{\ourmethod{}: Universal Relative Position Embedding across Geometric Spaces
} 

\titlerunning{\ourmethod{}}


\author{Yichen Xie\inst{1,2,}\textsuperscript{\#} \and
Depu Meng\inst{1} \and
Chensheng Peng\inst{1,2,}\textsuperscript{\#} \and
Yihan Hu\inst{1} \and
Quentin Herau\inst{1} \and 
Masayoshi Tomizuka\inst{2} \and 
Wei Zhan\inst{1,2}
}

\authorrunning{Y.~Xie et al.}

\institute{Applied Intuition \and University of California, Berkeley
}

\maketitle

\begingroup
\renewcommand{\thefootnote}{\#}
\footnotetext{Work done during internship at Applied Intuition.}
\endgroup

\begin{abstract}
Relative position embedding has become a standard mechanism for encoding positional information in Transformers. However, existing formulations are typically limited to a fixed geometric space, namely 1D sequences or regular 2D/3D grids, which restricts their applicability to many computer vision tasks that require geometric reasoning across camera views or between 2D and 3D spaces. To address this limitation, we propose \ourmethod{}, a universal extension of Rotary Position Embedding (RoPE) to cross-view or cross-dimensional geometric spaces. For each key/value image patch, \ourmethod{} samples 3D points along the corresponding camera ray at predefined depth anchors and projects them into the query image plane. Standard 2D RoPE can then be applied using the projected pixel coordinates. \ourmethod{} is a parameter-free and intrinsics-aware relative position embedding that is invariant to the choice of global coordinate systems, while remaining fully compatible with existing RoPE-optimized attention kernels. We evaluate \ourmethod{} as a plug-in positional encoding for transformer architectures across a diverse set of tasks, including novel view synthesis, 3D object detection, object tracking, and depth estimation, covering 2D–2D, 2D–3D, and temporal scenarios. Experiments show that \ourmethod{} consistently improves the performance of transformer-based models across all tasks, demonstrating its effectiveness and generality for geometric reasoning. Our code is available on our project website: \href{https://urope-pe.github.io/}{https://urope-pe.github.io/}.
\end{abstract}

\section{Introduction}
Transformers~\cite{vaswani2017attention} have
become the dominant architecture in the field of computer
vision including multi-view perception and generation tasks, from
novel view synthesis~\cite{jin2024lvsm,sajjadi2022scene} and stereo
matching~\cite{xu2023unifying,li2022practical}, to 2D/3D object
detection~\cite{carion2020end,meng2021conditional,liu2022petr,wang2023exploring}. A critical challenge
in applying Transformers to geometric tasks is how to encode
the spatial relationships between tokens that originate from
different viewpoints, coordinate systems, or even different
geometric modalities (2D images and 3D points).

As a standard solution, position embeddings inject positional information into the
permutation invariant Transformer architecture. Compared with absolute position embeddings~\cite{vaswani2017attention}, relative formulations~\cite{su2024roformer,shaw2018self,dai2019transformer,press2021train}
offer improved generalization and length extrapolation, making them a mainstream choice
in modern Transformers, especially for geometric tasks. A particularly
important development is Rotary Position Embedding
(RoPE)~\cite{su2024roformer}, which encodes \emph{relative}
rather than absolute positions through rotation matrices applied to
query/key pairs. 
Standard RoPE operates within a single flat coordinate space, assigning positions by
sequence indices (1D) or image grid locations (2D), which is fundamentally inadequate for
cross-view geometric reasoning: \textit{pixels from two camera views may be
close in 3D space yet far apart in their respective 2D grids.}

\begin{figure}[t]
    \centering
    \includegraphics[width=\textwidth]{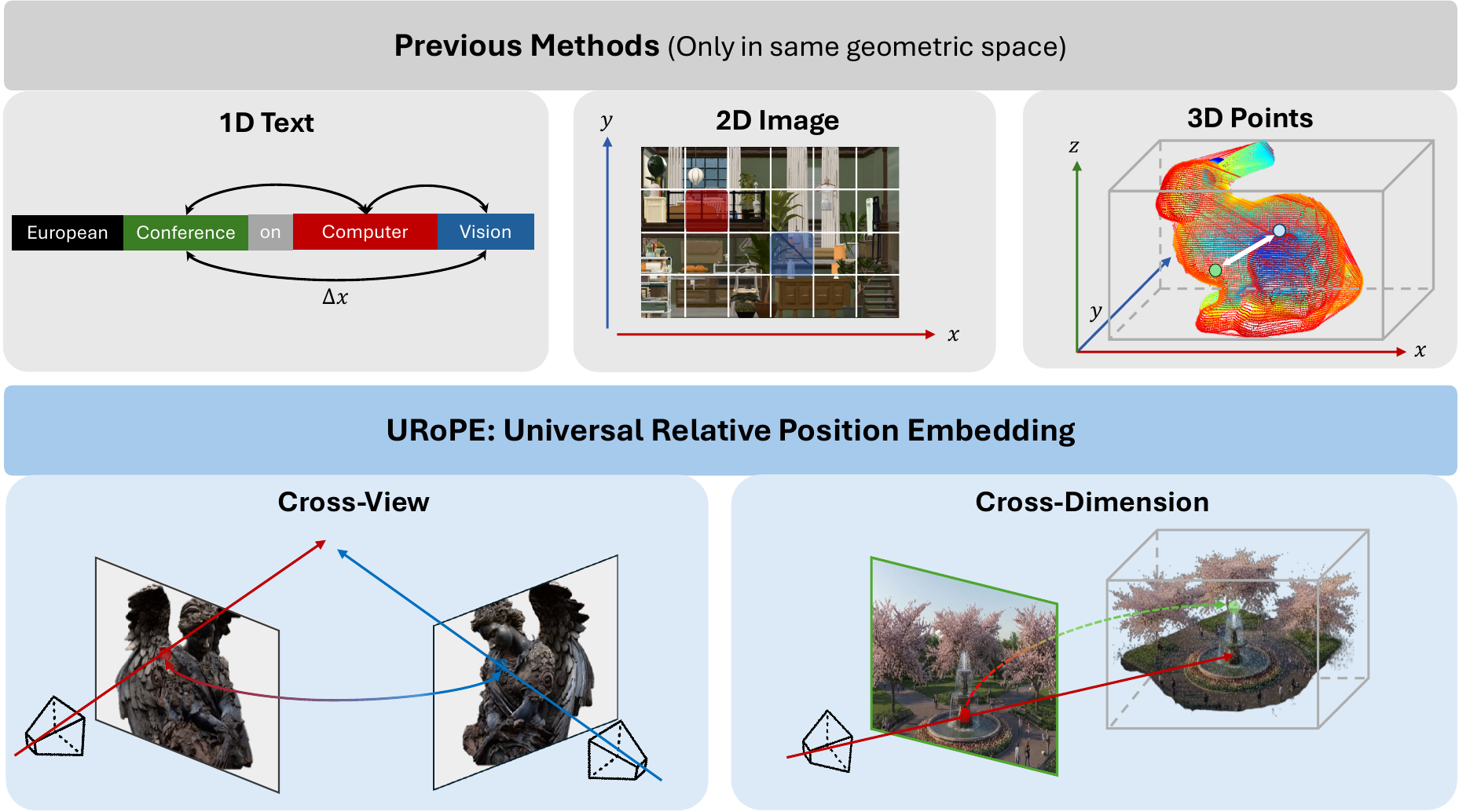}
    \caption{\ourmethod{} for Relative Position Embedding across Geometric Spaces. Previous relative position embeddings can only handle a shared geometric space (top), while \ourmethod{} focuses on the relative position embedding across geometric spaces (bottom).}
    \label{fig:teaser}
    \vspace{-10pt}
\end{figure}

Recent efforts have begun to incorporate camera geometry into attention,
but often rely on matrix multiplication rather than extending RoPE itself to cross-view settings.
Pl\"{u}cker ray embeddings~\cite{sitzmann2021light} concatenate ray origin and direction as input features. This is an absolute encoding that lacks the relative bias properties crucial for attention generalization, while its relative variant~\cite{xie2026raynova} still relies on the definition of a global coordinate system. 
Recent work~\cite{miyato2023gta,li2025cameras} combines camera parameters and standard 2D RoPE in a channel-wise split. However, the inter-camera geometry and intra-image spatial position are encoded disjointly with no interactions, and the inter-camera geometry is only considered for each camera rather than the local patch. Overall, it remains an open question to model the relative position across camera views and wider geometric spaces.

We observe that the fundamental question in cross-view relative position is:
\emph{where does a key token's 3D content appear in the query token's image?}
Instead of encoding abstract ray coordinates,
\ourmethod{} uses explicit projective geometry to express cross-view correspondences \emph{directly in the query image plane},
so the relative position can be modeled in a single shared coordinate system.

Specifically, for each key token from a source view, \ourmethod{} starts from its camera ray and
samples 3D points along the ray at a set of fixed depth anchors. Each sampled 3D point is then
projected into the query camera using the relative camera transform and the query intrinsics,
yielding a depth-conditioned pixel coordinate in the query image plane. Finally, we apply
standard 2D RoPE between the query location and the projected key location, producing a
geometry-aware \emph{relative} positional bias that is consistent across camera views.

A central challenge is that cross-view projection is inherently depth ambiguous: a source pixel
corresponds to an epipolar line in the query view. We address this with \textbf{depth-anchored multi-head attention}:
different attention heads (or head groups) are assigned with different fixed depth anchors, so each head
encodes one depth hypothesis and multi-head attention jointly covers near- to far-field correspondences.
In each head, we split the per-head channels across the horizontal and vertical axes ($d_h/2$ for $u$ and $d_h/2$ for $v$)
and apply standard 2D RoPE in the query image plane, keeping \ourmethod{} natively compatible with FlashAttention~\cite{dao2022flashattention}
and other RoPE-optimized kernels.

A key contribution of \ourmethod{} is its universality. We demonstrate that the same projective RoPE formulation
without task-specific modifications, serves as a plug-in position encoding across diverse geometric tasks:
\begin{itemize}
    \item \textbf{Novel view synthesis}: 2D$\rightarrow$2D cross-view attention with known camera poses on
    Objaverse~\cite{deitke2023objaverse} and RealEstate10k~\cite{zhou2018stereo}.
    \item \textbf{3D object detection and tracking}: 2D$\rightarrow$3D attention between image features and 3D query positions on
    nuScenes~\cite{caesar2020nuscenes}.
    \item \textbf{Stereo depth estimation}: 2D$\rightarrow$2D cross-view matching for depth prediction on
    RGBD~\cite{sturm2012benchmark}, Scenes11~\cite{ummenhofer2017demon}, and SUN3D~\cite{xiao2013sun3d}.
\end{itemize}

We show consistent improvements across all benchmarks, establishing \ourmethod{} as a general-purpose geometric position
encoding for Transformers.

\section{Related Work}

\begin{table}[t]
\centering
\caption{Comparison of Cross-View Position Embedding.
\textbf{Mechanism:} how geometry modulates attention: 
concatenation with input tokens (Concat.), 
matrix multiplication on Q/K/V features (MatMul), 
sinusoidal rotation on Q/K (RoPE), or a combination. 
\textbf{Per-patch Geo.:} whether the camera geometry is encoded at
the individual patch level. \textbf{SE(3) Inv.:} invariance to rigid
transformations of the global coordinate system. 
\textbf{Param-Free:} no learnable parameters introduced.}
\label{tab:comparison}
\vspace{-10pt}
\setlength{\tabcolsep}{6pt}
\resizebox{\textwidth}{!}{
\begin{tabular}{lllccc}
\toprule
\textbf{Method}  & \textbf{Mechanism} & \textbf{Geometric Info} & \textbf{Per-patch Geo.} & \textbf{SE(3) Inv.} & \textbf{Param-Free} \\
\midrule
Pl\"ucker~\cite{sitzmann2021light} & Concat. & 6D Ray & \cmark & \xmark & \cmark \\
Relative Ray~\cite{xie2026raynova} & RoPE & 6D Ray  & \cmark & \xmark & \cmark\\
GTA~\cite{miyato2023gta} & MatMul & Extrinsics & \xmark & \cmark & \cmark \\
P-RoPE~\cite{li2025cameras} & RoPE + MatMul & Proj. matrix + grid & \xmark  & \cmark & \cmark \\
RayRoPE~\cite{wu2026rayrope} & RoPE + Linear & Ray + learned depth & \cmark & \cmark & \xmark \\
\textbf{\ourmethod{}} & RoPE & Proj. coords & \cmark & \cmark & \cmark \\
\bottomrule
\end{tabular}
}
\vspace{-10pt}
\end{table}

\noindent\textbf{Position Encoding in Transformers.}
Since sinusoidal positional encodings are introduced for sequence modeling\cite{vaswani2017attention}, position embeddings have become standard
in transformers~\cite{dosovitskiy2020image,carion2020end} for multiple tasks. As an important milestone, Rotary Position Embedding
(RoPE)~\cite{su2024roformer} encodes relative positions through rotation
matrices applied to query and key vectors, enabling relative
position bias without explicit bias terms. RoPE has
become the default position encoding in modern large
language models~\cite{touvron2023llama,grattafiori2024llama}, which is also extended to 2D for
vision and multi-modal tasks~\cite{heo2024rotary,wang2024qwen2}.
Our work extends RoPE to cross-view and cross-dimensional
geometric spaces by utilizing the explicit projection to convert the key tokens to the same geometric space of query tokens.

\noindent\textbf{Transformers across Geometric Spaces.}  Transformers are widely applied to computer vision tasks across geometric spaces including novel view synthesis~\cite{hong2023lrm,jin2024lvsm}, 3D scene understanding~\cite{wang2022detr3d,liu2022petr,xie2023sparsefusion}, and stereo depth estimation~\cite{xu2023unifying}. 
To bridge the geometric gap, some method adopts explicit projection.
For example, epipolar attention~\cite{he2020epipolar,xie2024x,charatan2024pixelsplat} restricts attention to the projected epipolar lines but can not capture off-epipolar-line correspondences. DETR3D~\cite{wang2022detr3d} projects 3D queries to multiview image for feature sampling. BEVFormer~\cite{li2024bevformer} applies deformable attention~\cite{zhu2020deformable} with 3D reference points. However, these complex attention masks or kernels are unfriendly with efficient implementations like FlashAttention~\cite{dao2022flashattention}. In contrast, \ourmethod{} compute the RoPE positions via projective geometry,
natually encoding epipolar constraints through the soft attention bias.
By operating entirely through RoPE rather than matrix multiplications,
 \ourmethod{} unified inter-camera and geometry and intra-image spatial
position into a single mechanism while keeping the original Q/K/V multiplication format of standard attention operation.

\noindent\textbf{Multiview Position Embedding.}  Incorporating camera geometry into multi-view attention
can be broadly categorized by mechanism.
At the input level, Pl\"{u}cker ray embeddings~\cite{sitzmann2021light}
and camera ray maps~\cite{du2023learning}  concatenate ray information with
image features, while PETR~\cite{liu2022petr} and StreamPETR~\cite{wang2023exploring}
encodes 3D positions
into image features via position-aware embeddings.
These are absolute encodings
that lack the relative position bias property. RAYNOVA~\cite{xie2026raynova} takes a step further to encode the relative position in the ray space, but it still relies on the global coordinate system.  GTA~\cite{miyato2023gta} applies camera extrinsic matrices to Q/K/V features, achieving SE(3)-invariant. P-RoPE~\cite{li2025cameras} extends GTA by incorporating intrinsics via normalized projection matrices and combining with 2D RoPE for each patch positions. However, it decouples inter-camera and intra-camera geometry across separate head dimension blocks. RayRoPE~\cite{wu2026rayrope}, concurrent with our work, represents each patch as a ray segment with a layer-wise parametric module to predict the depth, but the unsupervised module offers no
guarantee to estimate the actual scene depth. In contrast, \ourmethod{} assigns fixed depth anchors across head groups, providing explicit multi-depth coverage without
learnable parameters.


\section{Method}

\begin{figure}[t]
    \centering
    \includegraphics[width=1.0\linewidth]{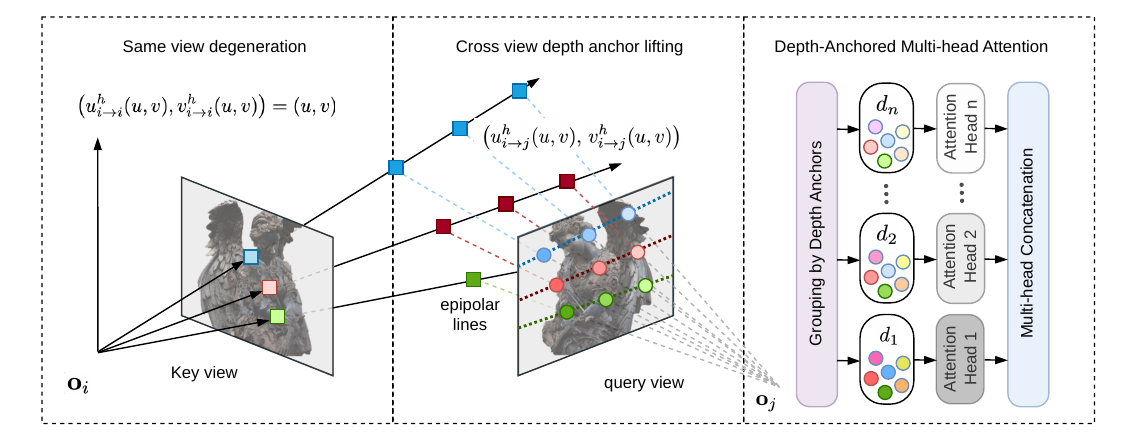}
    \vspace{-10pt}
    \caption{Overview of \ourmethod{}. \textbf{Left}: \ourmethod{} reduces to standard 2D RoPE in same-view case. \textbf{Middle}: for cross-view interaction, a key-view pixel is lifted along its camera ray to multiple anchored depths and projected into the query view, producing anchor-specific positions along epipolar lines. \textbf{Right}: projected positions are assigned to different attention heads according to depth anchors as depth-anchored multi-head attention.}
    \label{fig:method}
    \vspace{-10pt}
\end{figure}

We introduce \ourmethod{}, a relative position embedding for cross-view and cross-dimensional geometric reasoning. Our key insight is to use projective geometry to express cross-view correspondences directly in the query image plane. We first formalize camera rays as a bridge between 2D image coordinates and 3D space (Sec.~\ref{sec:ray}), then derive depth-anchored lifting and cross-view projection (Sec.~\ref{sec:projection}), and finally integrate the resulting projected coordinates into RoPE-based attention (Sec.~\ref{sec:transformer}). The overview of our method is visualized in Fig.~\ref{fig:method}.

\subsection{Camera Rays as a Bridge Between 2D and 3D}
\label{sec:ray}
Each image pixel corresponds to a light ray passing through the camera center. Camera rays therefore provide a natural bridge between the 2D image plane and 3D space. Given an image location $(u,v)$ in camera view $i$ with intrinsics $\mathbf{K}_i$ and extrinsics $[\mathbf{R}_i,\mathbf{t}_i]$, we represent its camera ray in the world coordinate system as
\begin{equation}
    \mathbf{ray}_i(u,v)=\left(\mathbf{o}_i,\mathbf{r}_i(u,v)\right), \quad
    \mathbf{r}_i(u,v)=\mathbf{R}_i^T\mathbf{K}_i^{-1}[u,v,1]^T,
\end{equation}
where $\mathbf{o}_i\in\mathbb{R}^{3}$ is the camera center in world coordinates, defined as $\mathbf{o}_i=-\mathbf{R}_i^T\mathbf{t}_i$, and $\mathbf{r}_i(u,v)\in\mathbb{R}^{3}$ is the corresponding ray direction.

The camera rays work as a natural absolute position embedding~\cite{jin2024lvsm} for attention modules across views. We also consider a direct ray-level relative position embedding baseline similar to~\cite{xie2026raynova} that applies independent 1D RoPE to the ray components $\mathbf{ray}_i(u,v)=(\mathbf{o}_i,\mathbf{r}_i(u,v))\in\mathbb{R}^6$ by splitting the per-head channels into six blocks (6D RoPE in Tab.~\ref{tab:nvs}).

However, the $\mathbf{ray}_i(u,v)$ representation can hardly model the potential intersection or spatial relationship between camera rays in the 3D space. Cross-view reasoning ultimately requires understanding where content observed along a ray in one camera would appear in another camera.  This motivates lifting points along the ray and explicitly projecting them into the query image plane.

\subsection{Bridging Geometric Gaps with Lifting and Projection}
\label{sec:projection}
To capture cross-view correspondences implied by camera rays, we consider 3D points along each ray. In the ideal case, a 3D point $\mathbf{p}_i(u,v)\in\mathbb{R}^{3}$ can be recovered from the ray given its depth. However, true depth is typically unavailable, and predicting depth inside every attention layer is undesirable.

\subsubsection{Depth-anchored lifting.}
We therefore introduce a set of fixed depth anchors $\mathcal{D}=\{d^h\}_{h=1}^K$ and lift each pixel $(u,v)$ into a set of 3D points
\begin{equation}
    \mathbf{p}_i^h(u,v) = \mathbf{o}_i + d^h \cdot \mathbf{r}_i(u,v),
    \label{eq:lifting}
\end{equation}
forming $\mathcal{P}_i(u,v)=\{\mathbf{p}_i^h(u,v)\}_{h=1}^K$.

\subsubsection{Projection across camera views.}
Given a source view $i$ and a query view $j$, we project each lifted point $\mathbf{p}_i^h(u,v)$ into the image plane of camera $j$:
\begin{align}
    \tilde{\mathbf{u}}_{i\rightarrow j}^h(u,v)
    &= \mathbf{K}_j\left(\mathbf{R}_j\mathbf{p}_i^h(u,v) + \mathbf{t}_j\right)
    = \begin{bmatrix}
        \tilde{u}_{i\rightarrow j}^h(u,v)\\
        \tilde{v}_{i\rightarrow j}^h(u,v)\\
        \tilde{w}_{i\rightarrow j}^h(u,v)
    \end{bmatrix}, \\
    u_{i\rightarrow j}^h(u,v) &= \tilde{u}_{i\rightarrow j}^h(u,v)\,/\,\tilde{w}_{i\rightarrow j}^h(u,v), \qquad
    v_{i\rightarrow j}^h(u,v) = \tilde{v}_{i\rightarrow j}^h(u,v)\,/\,\tilde{w}_{i\rightarrow j}^h(u,v),
    \label{eq:projection}
\end{align}
yielding a set of projected pixel coordinates $\{(u_{i\rightarrow j}^h(u,v), v_{i\rightarrow j}^h(u,v))\}_{h=1}^K$ in the query image plane. These projected points lie along the epipolar line induced by the source pixel, and provide an explicit, intrinsics-aware mapping from the source view to the query view.

\subsubsection{Degeneration to the single-view case.}
When $i=j$, projection reduces to the identity mapping, so the projected coordinates coincide with original locations:
\begin{equation}
\left(u_{i\rightarrow i}^h(u,v),v_{i\rightarrow i}^h(u,v)\right)=(u,v),\quad \forall h=1,\dots,K.
\end{equation}
Thus, \ourmethod{} naturally degenerates to standard 2D RoPE in a single image.

\subsubsection{Extension to 2D--3D interaction.}
\ourmethod{} also extends to 2D--3D interactions by skipping the image-plane projection. Given a 3D token at $(x,y,z)$, we measure relative positions between $(x,y,z)$ and the lifted points $\mathcal{P}_i(u,v)$ in 3D, enabling cross-attention between image features and 3D queries.

\subsection{Depth-Anchored Multi-head Attention}
\label{sec:transformer}

\subsubsection{\ourmethod{} for relative position embedding.}
We integrate the projected geometry derived in Sec.~\ref{sec:projection} into Transformer multihead attention. For self- or cross-attention across camera views, we consider a key/value image patch $(u,v)$ from camera view $i$ and a query image patch $(u',v')$ from camera view $j$. Let $h\in\{1,\dots,K\}$ denote the attention head index. We associate each head with a fixed depth anchor $d^{h}\in\mathcal{D}$. For head $h$, we lift the key/value patch using Eq.~\ref{eq:lifting} and project it into the query view using Eq.~\ref{eq:projection}, yielding
\begin{equation}
    \big(u^{h}_{i\rightarrow j}(u,v),\,v^{h}_{i\rightarrow j}(u,v)\big).
\end{equation}
Revealed in Sec.~\ref{sec:collapse}, this head-wise assignment would not cause the multi-head collapse, and the head-wise attention distribution are still balanced.

\subsubsection{Image-plane RoPE.}
We apply RoPE~\cite{su2024roformer} in the query image plane between the query location $(u',v')$ and the projected key location $\big(u^{h}_{i\rightarrow j}(u,v),v^{h}_{i\rightarrow j}(u,v)\big)$. Concretely, we apply standard 1D RoPE independently along the $x$- and $y$-axes on the first half of the per-head channels:

\begin{equation}
\begin{aligned}
    \mathbf{R}^{\text{query}}
    &=
    \text{diag}\!\left(
        \text{RoPE}_{\frac{C}{2}}(u'),\;
        \text{RoPE}_{\frac{C}{2}}(v')
    \right), \\
    \mathbf{R}^{\text{key}}(h)
    &=
    \text{diag}\!\left(
        \text{RoPE}_{\frac{C}{2}}\!\big(u^{h}_{i\rightarrow j}(u,v)\big),\;
        \text{RoPE}_{\frac{C}{2}}\!\big(v^{h}_{i\rightarrow j}(u,v)\big)
    \right),
\end{aligned}
\label{eq:urope}
\end{equation}
where $\text{RoPE}_{\frac{C}{2}}(\cdot)\in\mathbb{R}^{\frac{C}{2}\times \frac{C}{2}}$ is the 1D rotary position embedding~\cite{su2024roformer} and $C$ is the per-head dimension. 
Finally, we apply $\mathbf{R}^{\text{query}}$ to the query features and $\mathbf{R}^{\text{key}}(h)$ to each head of key features. This keeps \ourmethod{} fully compatible with RoPE-optimized attention kernels.


\subsubsection{Applying \ourmethod{} to multiview features.}
Although Eq.~\ref{eq:urope} resembles standard RoPE~\cite{su2024roformer}, it is not directly applicable when query tokens come from multiple camera views within the same sequence, since $\mathbf{R}^{\text{query}}_{\text{loc}}$ depends on the query view. Without loss of generality, we describe the self-attention case. Let $\mathbf{Q},\mathbf{K},\mathbf{V}\in\mathbb{R}^{B\times L\times H\times C}$ be the query/key/value tensors spanning $N$ views, where $L=N\cdot L_v$ and $L_v$ is the number of patches per view.

To ensure that all queries within one attention call share the same view, we reshape the queries by moving the view dimension from the sequence length into the batch dimension:
\begin{equation}
    \tilde{\mathbf{Q}}\in\mathbb{R}^{(B\cdot N)\times L_v\times H\times C}.
\end{equation}
Correspondingly, we repeat keys and values $N$ times along the batch dimension:
\begin{equation}
    \tilde{\mathbf{K}},\,\tilde{\mathbf{V}}\in\mathbb{R}^{(B\cdot N)\times L\times H\times C}.
\end{equation}
We then compute attention to obtain $\tilde{\mathbf{O}}\in\mathbb{R}^{(B\cdot N)\times L_v\times H\times C}$, and reshape it back to $\mathbf{O}\in\mathbb{R}^{B\times L\times H\times C}$. This rearrangement guarantees that each sample in $\tilde{\mathbf{Q}}$ corresponds to a single query view, enabling \ourmethod{} to be applied in the same manner as standard RoPE.

\subsubsection{Analysis of computation complexity.}
The time complexity of attention operation over $\tilde{\mathbf{Q}},\tilde{\mathbf{K}},\tilde{\mathbf{V}}$ is
\begin{equation}
    O(BN\times H\times L_v\times L\times C)=O(BHL^2C),
\end{equation}
which matches the complexity of attention over $\mathbf{Q},\mathbf{K},\mathbf{V}$. In terms of memory, the main overhead comes from repeating $\mathbf{K},\mathbf{V}$ along the batch dimension during the current layer's computation; this is transient and typically modest relative to the total memory footprint across layers. Overall, \ourmethod{} introduces no additional asymptotic computational burden to Transformers.

\section{Experiments}

In this section, we conduct extensive experiments to evaluate \ourmethod{} on three tasks including novel view synthesis (Sec~\ref{sec:novel-view}), 3D object detection (Sec~\ref{sec:3d-det}), and stereo depth estimation (Sec~\ref{sec:stereo}). For all the tasks, \ourmethod{} is integrated as a plug-in relative position embedding for the attention layers without any task- or model-specific designs. Afterwards, we give in-depth analysis of the role of \ourmethod{} in Sec.~\ref{sec:analysis}. Finally, we justify several key design modules in Sec.~\ref{sec:ablation}.

\subsection{Novel View Synthesis Task}
\label{sec:novel-view}
Novel view synthesis requires a comprehensive understanding of the 3D environment by aggregating visual information across multiple camera views. 

\noindent\textbf{Setup.} We combine \ourmethod{} with LVSM~\cite{jin2024lvsm} framework, a decoder-only transformer that performs self-attention across reference and target
view tokens. \ourmethod{} serves as the relative position embedding, replacing the pl\"ucker ray. We follow P-RoPE~\cite{li2025cameras} to reduce the model size due to limited computational resources. We conduct experiments on two datasets: Objaverse~\cite{deitke2023objaverse} and \\RealEstate10k~\cite{zhou2018stereo}. Objaverse~\cite{deitke2023objaverse} contains synthetic 3D objects, and the images are rendered from diverse viewpoints. To improve the difficulty, we randomly change the camera focal lengths for the images. RealEstate10k~\cite{zhou2018stereo} contains real-world indoor and
outdoor scenes gathered from YouTube videos. We evaluate the quality of novel view synthesis with three standard metrics: PSNR, SSIM, and LPIPS.

\noindent\textbf{Baselines.} We compare \ourmethod{} with \textbf{1) Pl\"ucker ray}: the absolute position embedding for camera rays adopted in LVSM~\cite{jin2024lvsm}; \textbf{2) 6D RoPE}: rotary position embedding in 6D ray space mentioned in Sec.~\ref{sec:ray}; \textbf{3) P-RoPE~\cite{li2025cameras}}: relative position embedding that combines RoPE on image space and camera parameters across views; \textbf{4) RayRoPE}~\cite{wu2026rayrope}: a concurrent work that attaches extra depth prediction module in attention layers. For fairness, we run all the baselines on the same datasets by ourselves.

\begin{figure}[tb!]
    \centering
    \includegraphics[width=\linewidth]{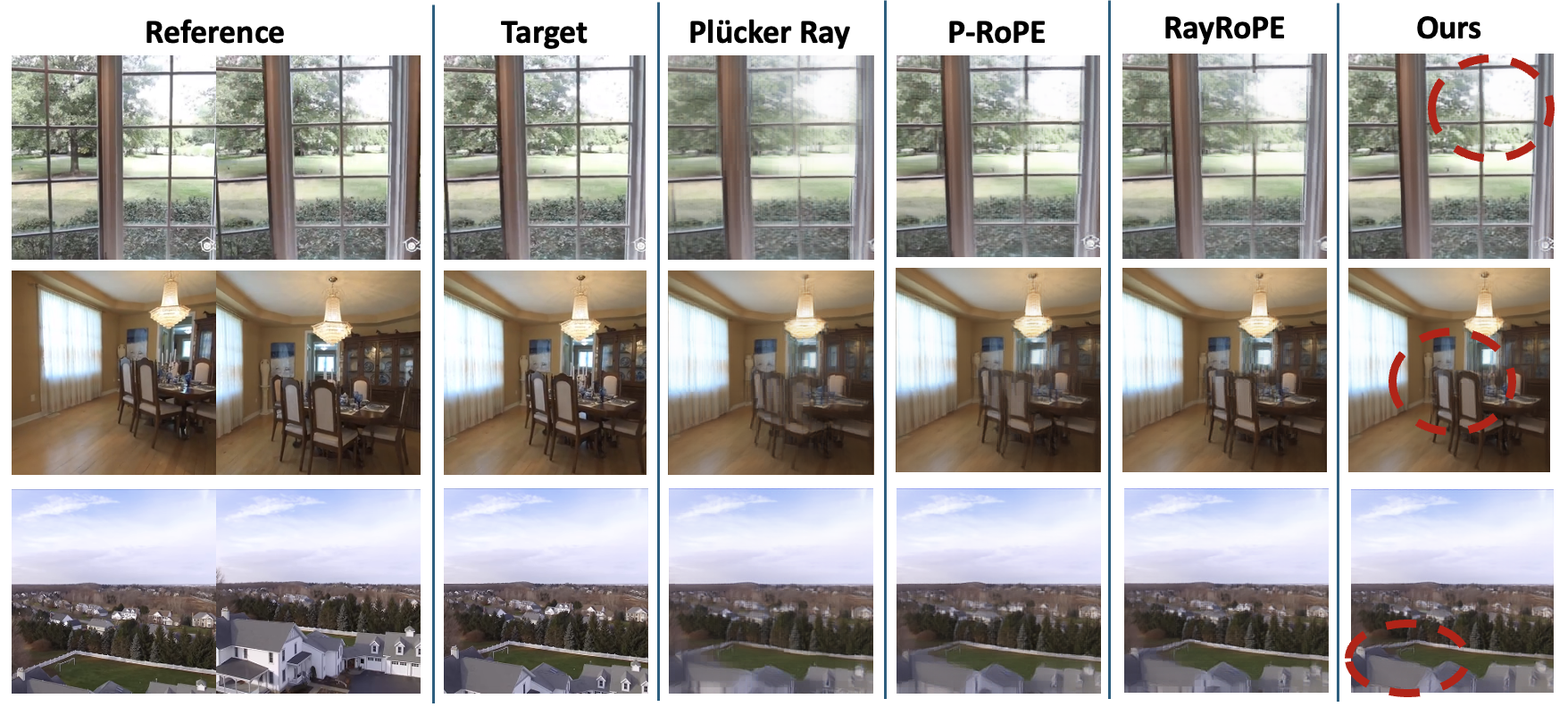}
    \vspace{-10pt}
    \caption{Qualitative Results for Novel View Synthesis. \ourmethod{} exploits relevant local information across views to synthesize sharper details.}
    \label{fig:nvs}
    \vspace{-10pt}
\end{figure}

\noindent\textbf{Results.} We report the quantitative comparison between different methods in Tab.~\ref{tab:nvs}. On both datasets, a notable performance gain of \ourmethod{} is witnessed compared to all the baselines. On the Objaverse dataset, \ourmethod{} can handle the challenging case of different camera intrinsic parameters, thanks to the design of explicit projection. On the RealEstate10k dataset, \ourmethod{} also performs well on large-scale scenes with complex geometry and appearance. The results show that \ourmethod{} can handle the self-attention including intra- and inter-view image features. By using fixed depth anchors, \ourmethod{} aligns cross-view features without additional parametric depth modules, simplifying optimization and avoiding errors from inaccurate depth prediction. Fig.~\ref{fig:nvs} shows qualitative examples, where \ourmethod{} improves LVSM’s ability to render fine-grained details.

\begin{table*}[tb!]
    \centering
    \caption{Results on Novel View Synthesis}
    \label{tab:nvs}
    \vspace{-10pt}
    \begin{tabular}{c|ccc|ccc}
        \toprule
        \multirow{2}{*}{\textbf{Methods}} & \multicolumn{3}{c|}{\textit{Objaverse}} & \multicolumn{3}{c}{\textit{RealEstate10k}}\\
        & PSNR$\uparrow$ & SSIM$\uparrow$ & LPIPS$\downarrow$ & PSNR$\uparrow$ & SSIM$\uparrow$ & LPIPS$\downarrow$\\
        \midrule
        Pl\"ucker Ray~\cite{jin2024lvsm} &  22.28 & 0.856 & 0.279 & 23.95 & 0.764 & 0.118\\
        6D RoPE &24.42 & 0.891 & 0.191  &  25.73 & 0.819 & 0.086\\
        \midrule
        P-RoPE~\cite{li2025cameras} & 24.88 & 0.896 & 0.176 &  25.28 & 0.806 & 0.092\\
        RayRoPE~\cite{wu2026rayrope} &  24.96 & 0.897 & 0.175 & 24.94 & 0.799 & 0.097\\
        \midrule 
        \ourmethod{} (ours) & \textbf{25.09} & \textbf{0.900} & \textbf{0.165} & \textbf{26.02} & \textbf{0.827} & \textbf{0.080} \\
        \bottomrule
    \end{tabular}
\end{table*}

\subsection{3D Object Detection and Tracking Task}
\label{sec:3d-det}
Vision-based 3D perception aims to recognize objects in 3D space using 2D images. Transformer-based algorithms often depend on the cross-attention module for the interaction between 3D queries and image features. As mentioned at the end of Sec.~\ref{sec:projection}, \ourmethod{} can be generalized to this cross-dimensional interaction.

\noindent\textbf{Setup.} To show the compatibility of \ourmethod{}, we consider both single-frame and long-horizontal multi-frame algorithms for 3D object detection, namely PETR~\cite{liu2022petr} and StreamPETR~\cite{wang2023exploring}, with additional tracking module from CenterPoint~\cite{yin2021center}. The experiments are conducted on nuScenes dataset~\cite{caesar2020nuscenes}, which contains large-scale outdoor scenes with dynamic objects. We report the NDS and mAP metrics for object detection, and AMOTA metric for object tracking.

\begin{figure}[tb!]
    \centering
    \includegraphics[width=\linewidth]{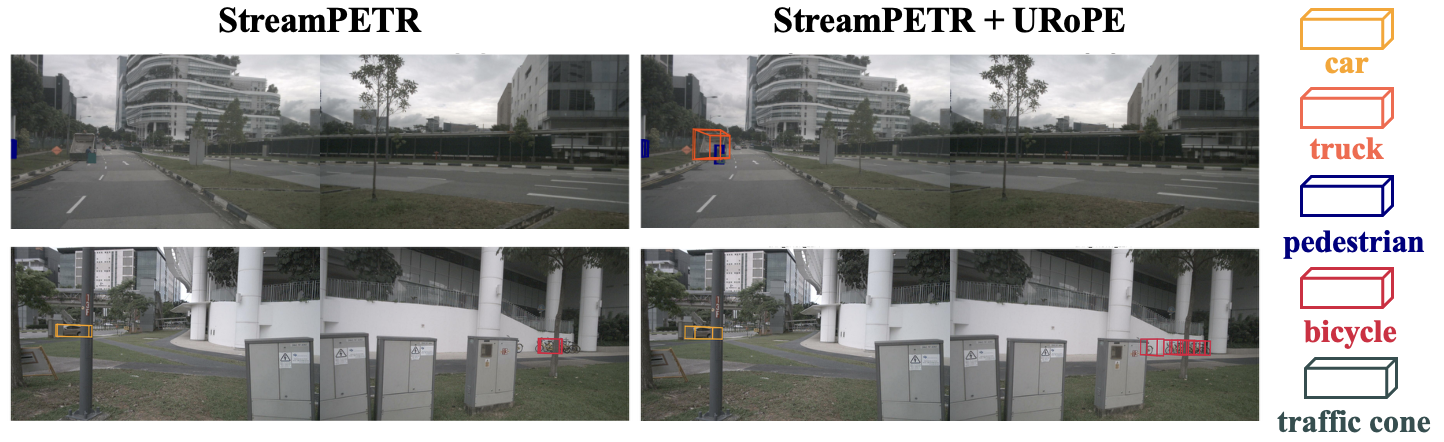}
    \vspace{-20pt}
    \caption{Qualitative Comparison on 3D Object Detection. \ourmethod{} can help to identify small details from multiframe and multiview images.}
    \label{fig:detection}
    \vspace{-10pt}
\end{figure}

\noindent\textbf{Results.} As reported in Tab.~\ref{tab:detection}, \ourmethod{} consistently improves 3D object detection and tracking performance, which verifies the effectiveness of \ourmethod{} in 2D-3D cross-dimensional interaction. \ourmethod{} helps to lift the visual features from 2D image space into 3D space by explicitly building correspondence between 3D queries and 2D image features. Especially, Tab.~\ref{tab:detection_temporal} shows that \ourmethod{} can handle not only multi-view camera images at the same time step but also long-horizontal multi-frame sequences with fast ego-camera motions. As demonstrated in Fig.~\ref{fig:detection}, \ourmethod{} improves the 3D object detection algorithm in the ability to identify small objects. Besides, the performance gain in tracking further demonstrates that \ourmethod{} enhances the robust and coherent recognition of objects along the entire trajectories at multiple frames.

\begin{table}[tb!]
    \centering
    \caption{Results on 3D Object Detection and Tracking}
    \label{tab:detection}
    \vspace{-20pt}
    \begin{subtable}{0.46\linewidth}
        \centering
        \caption{Single-Frame Multiview Performance}
        \vspace{-10pt}
    \resizebox{\linewidth}{!}{
        \begin{tabular}{c|cc|c}
        \toprule
            \multirow{2}{*}{\textbf{Methods}} & \multicolumn{2}{c|}{\textit{Detection}} & \textit{Tracking}\\
            & NDS$\uparrow$ & mAP$\uparrow$ & AMOTA$\uparrow$\\
            \midrule
            PETR~\cite{liu2022petr} & 34.9 & 30.9 & 0.222\\
            PETR + \ourmethod{} & \textbf{37.3} & \textbf{32.2} & \textbf{0.255}\\
        \bottomrule
        \end{tabular}
    }
    \end{subtable}
    \hfill
    \begin{subtable}{0.53\linewidth}
        \centering
        \caption{Multi-Frame Multiview Performance}
        \label{tab:detection_temporal}
        \vspace{-10pt}
    \resizebox{\linewidth}{!}{
        \begin{tabular}{c|cc|c}
        \toprule
            \multirow{2}{*}{\textbf{Methods}} & \multicolumn{2}{c|}{\textit{Detection}} & \textit{Tracking}\\
            & NDS$\uparrow$ & mAP$\uparrow$ & AMOTA$\uparrow$\\
            \midrule
            StreamPETR~\cite{wang2023exploring} & 47.6 & 37.5 & 0.335\\
            StreamPETR + \ourmethod{} & \textbf{50.6} & \textbf{41.1} & \textbf{0.380}\\
        \bottomrule
        \end{tabular}
    }
    \end{subtable}
\end{table}

\subsection{Stereo Depth Estimation Task}
\label{sec:stereo}
Stereo depth estimation exploits parallax to predict depth, which requires information transfer between two camera views. \ourmethod{} can also be integrated into stereo depth estimation algorithms, which can help correspond visual features across camera views.

\noindent\textbf{Setup.} We integrate \ourmethod{} into UniMatch~\cite{xu2023unifying}, which is a transformer-based method for stereo depth estimation. We conduct experiments on three datasets: RGBD~\cite{sturm2012benchmark}, SUN3D~\cite{xiao2013sun3d}, and Scenes11\cite{ummenhofer2017demon}, where RGBD and SUN3D datasets contain real scenes, but Scenes11 dataset is composed with synthetic scenes. We evaluate the performance with standard metrics for depth estimation task including Abs Rel, Sq Rel, RMSE, and RMSE log.

\noindent\textbf{Results.} We report the metrics on three datasets in Tab.~\ref{tab:depth}. For stereo depth estimation, there are only two camera views with a small distance between each other, and the model has a decoupled design for intra-view self-attention and inter-view cross-attention. Both factors simplify the combination of multiview visual information, and weaken the advantage of explicit projection. However, the model can still benefit from \ourmethod{} to consistently improve its performance in this simple case, which reflects the versatility of our proposed method.

\begin{table}[tb!]
    \centering
    \caption{Results on Stereo Depth Estimation. $^{\dag}:$ The
“Sq Rel” metric is less reliable on the RGBD dataset due to the imperfect depth and camera pose~\cite{ummenhofer2017demon}.}
    \label{tab:depth}
    \vspace{-5pt}
    \begin{tabular}{l|c|cccc}
    \toprule
      \textbf{Datasets} & \textbf{Methods} &  Abs Rel$\downarrow$ & Sq Rel$\downarrow$ & RMSE$\downarrow$ & RMSE log$\downarrow$\\
      \midrule
      \multirow{3}{*}{\textit{RGBD~\cite{sturm2012benchmark}}} & UniMatch~\cite{xu2023unifying} & 0.123&0.175$^{\dag}$ & 0.678&0.203   \\
      & UniMatch + P-RoPE~\cite{li2025cameras} & 0.105 & 0.203$^{\dag}$& 0.573 & \textbf{0.181}\\
      & UniMatch + \ourmethod{} & \textbf{0.103} & 0.201$^{\dag}$ & \textbf{0.571} & \textbf{0.181}\\
      \midrule
      \multirow{3}{*}{\textit{Scenes11~\cite{ummenhofer2017demon}}} & UniMatch~\cite{xu2023unifying} &  0.065 & 0.085 & 0.575 & 0.126 \\
      & UniMatch + P-RoPE~\cite{li2025cameras} & \textbf{0.049} & 0.063 & 0.474 & \textbf{0.104}\\
      & UniMatch + \ourmethod{} & \textbf{0.049} & \textbf{0.062} & \textbf{0.450} & \textbf{0.104}\\
      \midrule
      \multirow{3}{*}{\textit{SUN3D~\cite{xiao2013sun3d}}}  & UniMatch~\cite{xu2023unifying} & 0.131 & 0.098 & 0.397 & 0.169\\
      & UniMatch + P-RoPE~\cite{li2025cameras} & 0.117 & 0.075 & 0.343 & 0.152 \\
      & UniMatch + \ourmethod{} & \textbf{0.112} & \textbf{0.063} & \textbf{0.329} & \textbf{0.148}\\
    \bottomrule
    \end{tabular}
\end{table}

\begin{figure}[tb]
    \centering
    \begin{subfigure}{0.48\linewidth}
        \includegraphics[width=\textwidth]{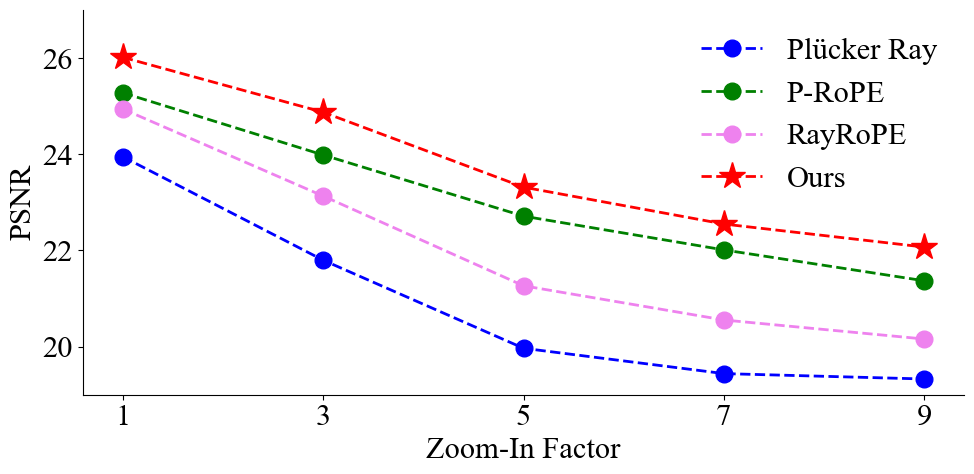}
        \caption{Different Camera Focal Lengths.}
    \end{subfigure}
    \hfill
    \begin{subfigure}{0.48\linewidth}
        \includegraphics[width=\textwidth]{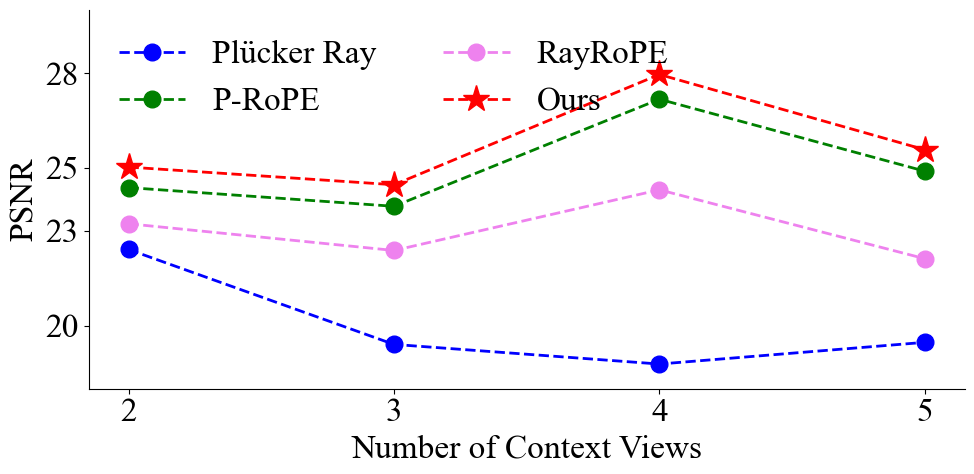}
        \caption{Different Context View Numbers.}
        \label{fig:robust_view}
    \end{subfigure}
    \vspace{-5pt}
    \caption{Robustness Analysis. Although the model is trained with fixed camera focal length and 2 context views, \ourmethod{} can maintain a robust performance in out-of-distribution cases. Fig.~\ref{fig:robust_view} adopts different context view selection with Tab.~\ref{tab:nvs}, so the metrics are not aligned.}
    \label{fig:robustness}
    \vspace{-10pt}
\end{figure}

\subsection{Analysis}
\label{sec:analysis}

We delve into the mechanism of \ourmethod{} to boost model performance. Unless otherwise specified, all the experiments are conducted on novel view synthesis task using RealEstate10K~\cite{zhou2018stereo}, following the setting in Sec.~\ref{sec:novel-view}.

\noindent\textbf{Scaling-Up Experiments.}
As a relative position embedding, it is critical for \ourmethod{} to exhibit improved performance with scaled-up computational resources. To evaluate the scalability of \ourmethod{}, we scale up the computation by enlarging the training batch size, network depth, and network width, which yields a 50$\times$ larger computational cost. As shown in Tab.~\ref{tab:scalability}, \ourmethod{} still brings reasonable performance gains compared to the default setting with scaled-up computational resources reflecting the great scalability of our proposed \ourmethod{}.

\begin{table}[t]
    \centering
    \begin{minipage}[t]{0.48\linewidth}
    \centering
    \caption{Scaling-Up (50$\times$) Experiment}
    \label{tab:scalability}
    \vspace{-10pt}
    \begin{tabular}{c|ccc}
        \toprule
        \textbf{Methods} & PSNR$\uparrow$ & SSIM$\uparrow$ & LPIPS$\downarrow$\\
        \midrule
        Pl\"ucker Ray~\cite{jin2024lvsm} &  28.66 & 0.889 &  0.113 \\
        \midrule 
        \ourmethod{} (ours) & \textbf{29.24} & \textbf{0.897} & \textbf{0.104}\\
        \bottomrule
    \end{tabular}  
    \end{minipage}
    \hfill
    \begin{minipage}[t]{0.48\linewidth}
        \centering
        \caption{\ourmethod{} with Available Depths.}
        \label{tab:known-depth}
        \vspace{-10pt}
        \begin{tabular}{c|ccc}
        \toprule
       \textbf{Methods}  & PSNR$\uparrow$ & SSIM$\uparrow$ & LPIPS$\downarrow$ \\
        \midrule
        \ourmethod{} & 26.02 & 0.827 & 0.080\\
        \midrule
        \textit{+ context depths}& \textbf{26.31} & \textbf{0.832} & \textbf{0.078} \\
    \bottomrule
    \end{tabular}
    \end{minipage}
    \vspace{-5pt}
\end{table}

\subsubsection{Combination with Known Depths.}
\ourmethod{} can exploit available depth information. We label the depths of context views in RealEstate10K with DepthAnythingV3~\cite{lin2025depth}, while the target view depths are unavailable to avoid information leakage. For context views with available depth, we replace the fixed depth anchors with concrete depth values, while anchors for target views without depths remain unchanged. Tab.~\ref{tab:known-depth} shows that \ourmethod{} can benefit from available depth information to improve performance. 

\subsubsection{Robustness Analysis.} Firstly, we consider two out-of-distribution scenarios in Fig.~\ref{fig:robustness}: 1) Different camera focal lengths and 2) More context views. Although \ourmethod{} is trained with fixed camera intrinsic parameters and two context views, it is robust to all these changes during inference. Since \ourmethod{} exploits explicit epipolar projection to obtain the spatial correlation across camera views, it is not very sensitive to the change of camera parameters and view number. As a result, it naturally has great robustness in out-of-distribution situations. Afterwards, we also analyze the robustness of \ourmethod{} to noisy camera parameters. Although \ourmethod{} takes camera extrinsic and intrinsic parameters as inputs, it has reasonable robustness to noisy camera parameters. In Tab.~\ref{tab:cam-robustness}, we conduct an additional experiment where random noise is injected into the camera extrinsic and intrinsic parameters separately. Results show that \ourmethod{} consistently outperforms baselines under camera parameter perturbations.

\begin{table}[t]
    \centering
    \caption{Robustness to Noisy Camera Parameters}
    \label{tab:cam-robustness}
    \vspace{-10pt}
    \resizebox{\linewidth}{!}{
    \begin{tabular}{c|ccc|ccc|ccc}
    \toprule
       \multirow{2}{*}{\textbf{Methods}} & \multicolumn{3}{c|}{\textit{perturbed focal lengths}} & \multicolumn{3}{c}{\textit{perturbed rots}} & \multicolumn{3}{|c}{\textit{perturbed rots} \& \textit{trans}}\\
       & PSNR$\uparrow$ & SSIM$\uparrow$ & LPIPS$\downarrow$ & PSNR$\uparrow$ & SSIM$\uparrow$ & LPIPS$\downarrow$ & PSNR$\uparrow$ & SSIM$\uparrow$ & LPIPS$\downarrow$\\
    \midrule
        Pl\"ucker Ray~\cite{jin2024lvsm} & 23.89 & 0.762 & 0.117& 22.73 & 0.719 & 0.140 & 22.01 & 0.696 & 0.152\\
        P-RoPE~\cite{li2025cameras} & 25.04 & 0.800 & 0.093 & 23.46 & 0.744 & 0.121 & 22.17 & 0.704 & 0.144\\
        \ourmethod{} & \textbf{25.83} & \textbf{0.822} & \textbf{0.085} & \textbf{24.07} & \textbf{0.766} & \textbf{0.113} & \textbf{22.81} & \textbf{0.728} & \textbf{0.132}\\
    \bottomrule
    \end{tabular}
    }
\end{table}

\subsubsection{Region-wise Anchor Depth Importance.} 
\begin{figure}[tb]
    \centering
    \includegraphics[width=0.9\linewidth]{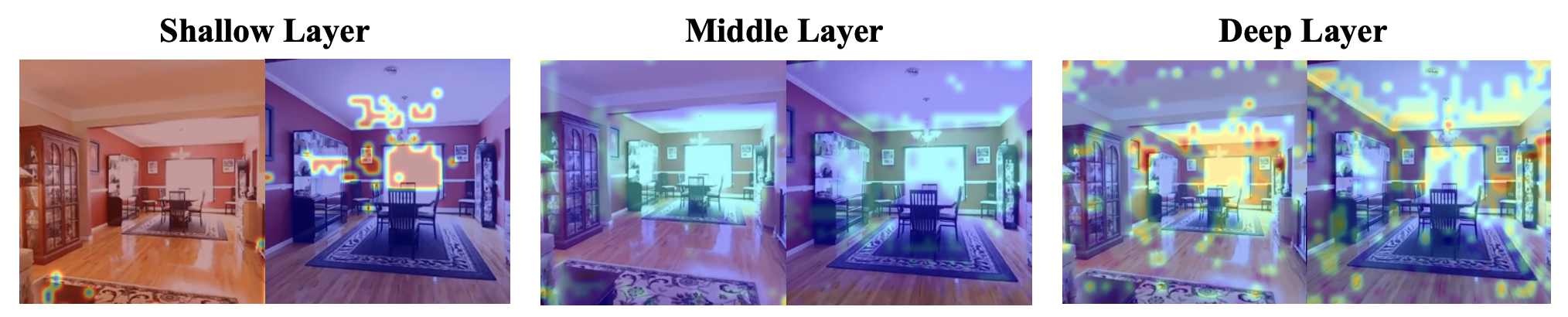}
    \vspace{-10pt}
    \caption{Region-wise Anchor Depth Importance. In the middle layer, the importance of depth anchors is highly related to the actual depth, while this correlation is not significant in both the shallow and deep layers.}
    \label{fig:anchor}
    \vspace{-5pt}
\end{figure}

Since we assign multiple anchor depths to each image patch, it is natural to think whether the actual depth affects the importance of each depth anchor. Since each depth anchor is associated with different attention heads, we can explore the head-wise importance in attention module for different regions as an indicator for the anchor depth importance. We visualize the distribution of the most important attention head in the self-attention for all the key/value image patches from context views, where the depth of anchors increases along with the order of the attention head. The strong correlation between real depth and anchor importance are only witnessed in the middle layer, as visualized in Fig.~\ref{fig:anchor}. This fact reveals that the network can learn to adaptively exploit different depth anchors based on the local information. In contrast, the correlation is weak in both the shallowest and deepest layers. In the shallow layers, the model does not have enough information to utilize specific anchor depths, while in the deep layers, the model may prefer learning more high-level semantic information beyond the depth value.

\begin{table}[tb!]
    \centering
    \caption{Analysis of Multihead Collapse. We calculate the attention score entropy across heads for all the key/value patches from the context views per attention layer.}
    \label{tab:collapse}
    \vspace{-10pt}
    \setlength{\tabcolsep}{6pt}
    \begin{tabular}{c|cccccc}
    \toprule
       \multirow{2}{*}{\textbf{Methods}}  & \multicolumn{6}{c}{Per-layer head-wise normalized entropy$\uparrow$}  \\
         & 1 & 2 & 3 & 4 & 5 & 6\\
    \midrule
        P-RoPE~\cite{li2025cameras} & 0.93 & 0.71 & 0.82 & 0.80 & 0.83 & 0.79\\
        \ourmethod{} & 0.95 & 0.87 & 0.84 & 0.80 & 0.82 & 0.83 \\
    \bottomrule
    \end{tabular}
    \vspace{-10pt}
\end{table}

\subsubsection{Risk of Multi-Head Collapse.}
\label{sec:collapse}
With head-wise depth-anchor assignment, a potential failure mode is \emph{multi-head collapse}, i.e., only a small subset of heads corresponding to specific depth anchors dominates attention for most key/value tokens. To examine this, we conduct a statistical analysis. For each key/value token from the context views, we compute its average attention score among all the query tokens from the target view for each head. Then, we calculate the entropy across all the attention heads, which is averaged over key/value tokens from multiple samples as a metric for multihead collapse and normalzed into $(0,1)$. Higher entropy indicates better balance across heads. In Tab.~\ref{tab:collapse}, we find the entropy values are similar between \ourmethod{} and P-RoPE~\cite{li2025cameras} across all the six layers. This shows that our head-wise depth anchor assignments would not cause the multihead collapse.

\subsection{Ablation Study}
\label{sec:ablation}
We justify the design of key modules in \ourmethod{} through ablation studies. The experiments are conducted on the RealEstate10k dataset~\cite{zhou2018stereo} using the LVSM model~\cite{jin2024lvsm} with the same setup as Sec.~\ref{sec:novel-view}. Due to different hyperparameters, the results may differ from Tab.~\ref{tab:nvs}, but the fairness of each comparison is guaranteed.

\begin{table}[tb!]
    \centering
    \caption{Effect of Combining Absolute and Relative Position Embeddings. The experiments are conducted on RealEstate10k using LVSM model.}
    \label{tab:ablation-absolute}
    \vspace{-10pt}
    \setlength{\tabcolsep}{4pt}
    \begin{tabular}{c|c|ccc}
    \toprule
        \textbf{\makecell[c]{Abs. Pos.}} & \textbf{\makecell[c]{Rel. Pos.}} & PSNR$\uparrow$ & SSIM$\uparrow$ & LPIPS$\downarrow$\\
        \midrule
         \textit{pl\"ucker ray} & - & 23.95 & 0.764& 0.118 \\ 
         \textit{pl\"ucker ray} & \textit{2D RoPE} & 24.96 & 0.795 & 0.098\\
         \midrule 
         \textit{pl\"ucker ray} & \textit{\ourmethod{}}& 25.89 & 0.826 & 0.078 \\
         \textit{local ray} & \textit{\ourmethod{}}& 26.00 & 0.828 & 0.077 \\
         - & \textit{\ourmethod{}}& 25.85 & 0.824 & 0.083\\
         \bottomrule
    \end{tabular}
    \vspace{-10pt}
\end{table}

\subsubsection{Combination of Absolute and Relative Position Embedding.} We explore whether \ourmethod{} can benefit from knowing the absolute camera ray position in the world coordinate by combining \ourmethod{} with the absolute position embedding. The results are reported in Tab.~\ref{tab:ablation-absolute}. \ourmethod{} can hardly achieve a performance gain when combined with global pl\"ucker rays since \ourmethod{} already contains enough information to reflect the spatial relationships across multi-view images. In contrast, \ourmethod{} can obtain some slight improvements if local ray directions in the camera coordinate are integrated since it reflects the intra-view positions. However, for simplicity, we do not include these local camera rays in other experiments by default.

\subsubsection{Depth Anchor Splitting.} We consider two ways to split anchor depth: head-wise \textit{v.s.} channel-wise split. Head-wise splitting assigns multiple anchor depths to different attention heads, while channel-wise splitting divides the per-head feature into different depth anchors. Due to the limited feature dimension, channel-wise splitting cannot support many anchor depths. As shown in Tab.~\ref{tab:ablation}, head-wise splitting can greatly improve performance. With head-wise splitting, we can accommodate more components of different frequencies to deal with both short-range and long-range relationships.

\subsubsection{Number of Depth Anchors.} With head-wise splitting, we consider different numbers of depth anchors. When the depth anchors are fewer than the attention heads, we divide the attention heads into groups, and each group shares the same depth anchor. In Tab.~\ref{tab:ablation}, we compare the performance with different numbers of depth anchors. Results show that the performance of \ourmethod{} is relatively robust for the anchor number. We think four anchor depths are already enough to reflect the position of the projected epipolar line, which helps to guide the attention module to focus on the most important regions across camera views.

\begin{table*}[tb!]
    \centering
    \caption{Ablation of \ourmethod{} on RealEstate10k with LVSM~\cite{jin2024lvsm}. \textbf{Number} is the count of fixed depth anchors. \textbf{Splitting} assigns anchors either to heads (head-wise) or to channels (channel-wise). \textbf{Sampling} controls how anchor depths are spaced (uniform, log-uniform, or LID~\cite{tang2020center3d}). \textbf{Depth} indicates whether we use fixed anchor depths or a learned per-layer depth prediction module.}
    \label{tab:ablation}
    \vspace{-10pt}
    \setlength{\tabcolsep}{4pt}
    \begin{tabular}{cccc|ccc}
        \toprule
        \textbf{\makecell[c]{Depth}} &
        \textbf{\makecell[c]{Number}} &
        \textbf{\makecell[c]{Splitting}} &
        \textbf{\makecell[c]{Sampling}} &
        PSNR$\uparrow$ & SSIM$\uparrow$ & LPIPS$\downarrow$\\
        \midrule
        \multirow{8}{*}{\textit{fixed anchors}}
        & 16 & \multirow{5}{*}{\textit{head-wise}} & \multirow{5}{*}{\textit{uniform}} & 25.85 & 0.824 & 0.083 \\
        & 8  &  &  & 25.84 & 0.823 & 0.084 \\
        & 4  &  &  & \textbf{26.01} & \textbf{0.827} & 0.082 \\
        & 2  &  &  & 25.94 & 0.826 & 0.082 \\
        & 1  &  &  & 25.37 & 0.807 & 0.092 \\
        \cmidrule(lr){2-7}
        & 2  & \textit{channel-wise} & \textit{uniform} & 25.47 & 0.815 & 0.087\\
        \cmidrule(lr){2-7}
        & \multirow{2}{*}{4} & \multirow{2}{*}{\textit{head-wise}} & \textit{LID} & \textbf{26.01} & \textbf{0.827} & \textbf{0.081} \\
        &    &  & \textit{log-uniform} & 25.94 & 0.825 & 0.082 \\
        \midrule
        \textit{learned} & -- & -- & -- & 25.57 & 0.818 & 0.085 \\
        \bottomrule
    \end{tabular}
    \vspace{-10pt}
\end{table*}

\subsubsection{Sampling of Depth Anchors.} We consider three different ways to sample depth anchors. 1) Uniform: The anchor depths are sampled uniformly within a range. 2) Log Uniform: The logarithm of anchor depths are sampled uniformly. 3) Linear-Increasing Discretization (LID)~\cite{tang2020center3d}: The bin size of each range linearly increases along the depth dimension. In Tab.~\ref{tab:ablation}, we compare these three different methods. We find that the uniform sampling and LID sampling achieve the similar great performance since both of them can represent the projected epipolar line in a balanced manner.

\subsubsection{Parametric Depth Estimation Module.} \ourmethod{} utilizes fixed depth anchors to build the relationship across the geometric gap. We consider an alternative way to predict the depth in each attention layer. In this case, we inject a lightweight parametric depth estimation module for each layer instead of the fixed depth anchors. As shown in Tab.~\ref{tab:ablation}, this parametric depth estimation brings worse performance than fixed depth anchors since it is challenging to predict accurate depth from the image feature, as it may not include enough geometric cues, especially in the shallow layers. In contrast, fixed anchors can lead to stable performance regardless of the semantics in the image feature.

\section{Conclusion}

In this paper, we presented \ourmethod{}, a universal extension of relative position embedding for cross-view and cross-dimensional geometric reasoning. By lifting key image tokens at fixed depth anchors and projecting them into the query image plane, \ourmethod{} enables standard 2D RoPE to encode cross-view correspondences in a shared image coordinate system. The resulting encoding is parameter-free, intrinsics-aware, invariant to global coordinate, and natively compatible with RoPE-optimized attention kernels. Experiments on novel view synthesis, 3D object detection, and stereo depth estimation show that \ourmethod{} consistently outperforms prior position encodings across all the benchmarks, while demonstrating strong out-of-distribution generalization. A current limitation is the reliance on known camera parameters. It is a promising direction to extend \ourmethod{} to uncalibrated settings in future work.

\FloatBarrier
\bibliographystyle{splncs04}
\bibliography{arxiv}

\newpage
\appendix
\renewcommand{\thefigure}{A.\arabic{figure}}
\renewcommand{\thetable}{A.\arabic{table}}

In this appendix, we exhibit additional experiment results in Sec.~\ref{sec:add_exp}. Afterwards, the implementation details of our experiments are explained in Sec.~\ref{sec:implementation}. Finally, we discuss some limitations of our work in Sec.~\ref{sec:limitation}.

\begin{table}[hb!]
    \centering
    \caption{Sensitivity to Depth Ranges.}
    \label{tab:range}
    \setlength{\tabcolsep}{6pt}
    \begin{tabular}{c|ccc}
    \toprule
        \textbf{Depth Range} & PSNR$\uparrow$ & SSIM$\uparrow$ & LPIPS$\downarrow$\\
        \midrule
         {[2m, 20m]} & \textbf{26.01} & \textbf{0.827} & 0.082  \\ 
         \midrule
         {[10m, 20m]} & 25.67 & 0.816 & 0.088  \\ 
         \midrule
         {[2m, 10m]} &25.93 & 0.826 & 0.082  \\ 
         {[2m, 30m]} & 25.98 & 0.826 & \textbf{0.081}\\
         {[2m, 40m]} & 25.88 & 0.824 & 0.083  \\ 
         \bottomrule
    \end{tabular}
\end{table}

\section{Additional Experiment Results}
\label{sec:add_exp}

\subsection{Sensitivity to Depth Ranges}
\ourmethod{} samples multiple depth anchors within a pre-defined depth range. We conduct an additional ablation study to evaluate the sensitivity of \ourmethod{} to the pre-defined depth ranges. All the experiment settings are same as Sec.~\ref{sec:ablation} and we sample 4 depth anchors within the range uniformly. We report the results in Tab.~\ref{tab:range}. It is important to cover the close regions, but \ourmethod{} is robust to different reasonable upper bounds. As a result, \ourmethod{} is not very sensitive to the selection of depth ranges.

\subsection{Camera-Controlled Video Generation}
\label{sec:video}
\begin{figure}[htb]
    \centering
    \begin{subfigure}{\linewidth}
        \includegraphics[width=\linewidth]{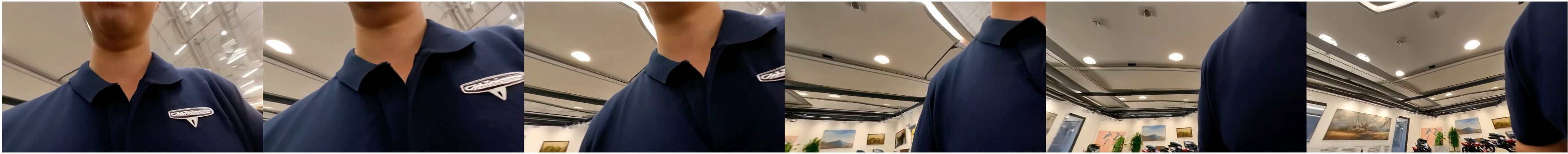}
        \caption{The given camera trajectory pans to the left.}
    \end{subfigure}
    \begin{subfigure}{\linewidth}
        \includegraphics[width=\linewidth]{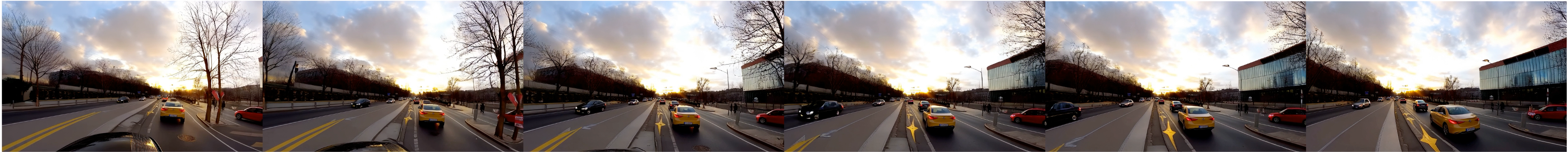}
        \caption{The given camera trajectory moves forward and pans to the right.}
    \end{subfigure}
    \caption{Qualitative Results for Camera-Controlled Video Generation}
    \label{fig:video}
\end{figure}

We integrate \ourmethod{} into Wan2.1-T2V-1.3B~\cite{wan2025wan} and finetune the attention modules. 
The final architecture is similar with UCPE~\cite{zhang2026unified}. 
In this case, the model can take camera trajectories as inputs to control the video generation. In Fig.~\ref{fig:video}, we provide some qualitative examples where the synthetic videos can follow the given per-frame camera movement.

\subsection{Additional Qualitative Results}
\begin{figure}[htb!]
    \centering
    \includegraphics[width=\linewidth]{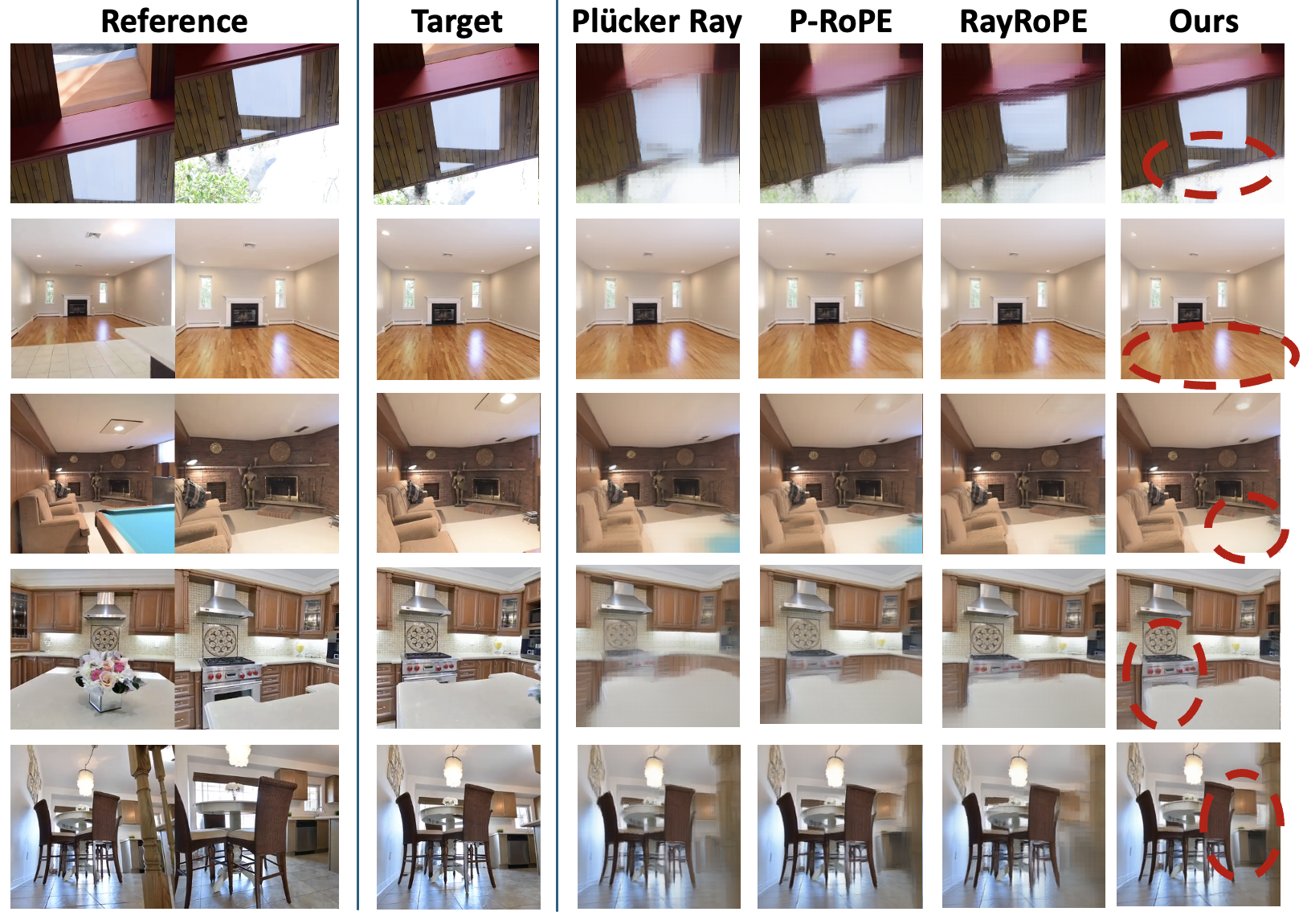}
    \caption{Additional Qualitative Results for Novel View Synthesis. \ourmethod{} helps to produce sharper details and synthetic novel views are more geometrically reasonable.}
    \label{fig:supp_nvs}
\end{figure}

\begin{figure}[htb!]
    \centering
    \includegraphics[width=\linewidth]{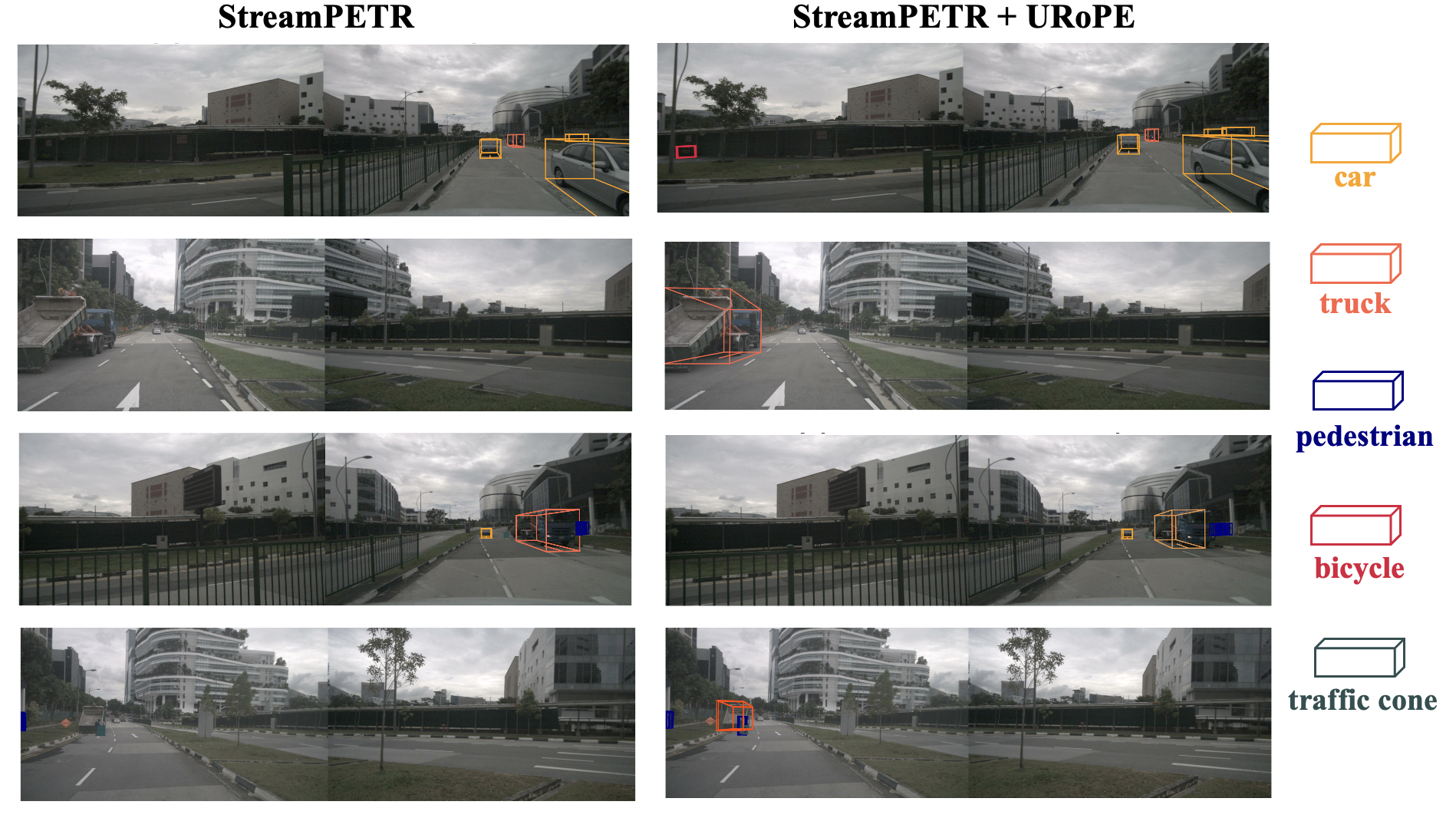}
    \caption{Additional Qualitative Results for 3D Object Detection. \ourmethod{} helps to identify some and difficult objects in the 3D space.}
    \label{fig:supp_det}
\end{figure}

In Fig.~\ref{fig:supp_nvs}, we visualize some additional results on novel view synthesis using different methods. Compared with other position embeddings, \ourmethod{} helps produce sharper details and the synthesized novel views are more spatially reasonable. In Fig.~\ref{fig:supp_det}, we also provide some additional visualizations for 3D object detection task. \ourmethod{} helps to identify some small and confusing objects.

\section{Experiment Details}
\label{sec:implementation}

\subsection{Novel View Synthesis Details}
We follow the implementation in \cite{li2025cameras} to reduce the computational burden of the original LVSM model~\cite{jin2024lvsm}. Some key implementation details are listed as follows.

\begin{itemize}
    \item The image resolution are fixed at $256\times256$ for both training and inference with a transformer patch size $8\times8$.
    \item Given the limited computational resources, we reduce the number of transformer blocks from 24 to 6. The MLP channel dimension is also reduced from 3072 to 1024. This modification allows us to train the model using 2 GPUs with a total batch size of 16 for 8k iterations on both RealEstate10k~\cite{zhou2018stereo} and Objaverse~\cite{deitke2023objaverse} datasets.
    \item For the Objaverse dataset~\cite{deitke2023objaverse}, we download the rendered images provided by Zero-1-to-3~\cite{liu2023zero}. To evaluate the robustness to camera parameters, we follow RayRoPE~\cite{wu2026rayrope} to randomly change the camera focal length with a random scale between 0.4 and 1.6.
\end{itemize}

To ensure the fair comparison, we reproduce all the methods in Tab.~\ref{tab:nvs} by ourselves on both datasets using their public codes.

\subsection{3D Object Detection and Tracking}
We integrate \ourmethod{} into PETR~\cite{liu2022petr} and StreamPETR~\cite{wang2023exploring} with their official code. Both models adopt ResNet50~\cite{he2016deep} backbone with a resolution of $256\times704$ for input images. They are trained for 24 epochs on nuScenes dataset~\cite{caesar2020nuscenes} with a total batch size of 16 using 2 GPUs. Both models include 900 queries that attend to multiview image features through 2D-3D cross-attention modules.

\subsection{Depth Estimation}
For depth estimation, we adopt UniMatch model~\cite{xu2023unifying}. It consists of intra-image self-attention and inter-image cross-attention modules. We follow \cite{li2025cameras} to integrate \ourmethod{} into both self-attention and cross-attention modules on the Q/K/V/O vectors. As explained in Sec.~\ref{sec:projection}, \ourmethod{} degrades to RoPE~\cite{heo2024rotary} in the intra-view self-attention. For all the three datasets, the model is trained for 100k steps using 3 GPUs with a total batch size of 78. The image resolution is set as $448\times576$.

\section{Limitations}
\label{sec:limitation}

\ourmethod{} relies on ground-truth camera intrinsic and extrinsic parameters for lifting and projection, which limits its applicability to tasks where camera calibration is available. Consequently, it cannot be directly applied to scenarios such as 3D reconstruction from uncalibrated camera observations. In addition, due to limited computational resources, we do not evaluate our approach on large-scale models. In future work, we plan to extend \ourmethod{} to operate with estimated camera parameters, enabling its integration with modern 3D reconstruction frameworks such as VGGT~\cite{wang2025vggt} and DepthAnythingV3~\cite{lin2025depth}.


\end{document}